\documentclass[reprint, cha]{revtex4-1}

\usepackage{amssymb}
\usepackage[english]{babel}    
\usepackage[dvips]{graphics}
\usepackage{subfigure}
\usepackage[pdftex]{graphicx}
\usepackage{multirow}
\usepackage{rotating}
\usepackage{picinpar}
\usepackage{color}
\usepackage{lineno}
\usepackage{listings}

\begin{document}

\title{Brachiaria species identification using imaging techniques based on fractal descriptors}

\author{Jo\~ao Batista Florindo}
 	     \email{jbflorindo@gmail.com}
\affiliation{Scientific Computing Group, S\~ao Carlos Institute of Physics, University of S\~{a}o Paulo (USP),  cx 369 13560-970 S\~{a}o Carlos, S\~{a}o Paulo, Brazil - www.scg.ifsc.usp.br} 

\author{N\'ubia Rosa da Silva}
 	     \email{nubiasrosa@gmail.com}
\affiliation{Institute of Mathematics and Computer Science, University of S\~{a}o Paulo (USP), Avenida Trabalhador s\~{a}o-carlense, 400 13566-590 S\~{a}o Carlos, S\~{a}o Paulo, Brazil} 

\author{Liliane Maria Romualdo}
 	     \email{lilianeromualdo@yahoo.com.br}
\affiliation{Faculdade de Zootecnia e Engenharia de Alimentos, Universidade de S\~{a}o Paulo (USP), Pirassunuga, S\~{a}o Paulo, Brazil} 

\author{Fernanda de Fátima da Silva}
 	     \email{ferdefatima@hotmail.com}
\affiliation{Faculdade de Zootecnia e Engenharia de Alimentos, Universidade de S\~{a}o Paulo (USP), Pirassunuga, S\~{a}o Paulo, Brazil} 

\author{Pedro Henrique de Cerqueira Luz}
 	     \email{phcerluz@usp.br}
\affiliation{Faculdade de Zootecnia e Engenharia de Alimentos, Universidade de S\~{a}o Paulo (USP), Pirassunuga, S\~{a}o Paulo, Brazil} 

\author{Valdo Rodrigues Herling}
 	     \email{vrherlin@usp.br}
\affiliation{Faculdade de Zootecnia e Engenharia de Alimentos, Universidade de S\~{a}o Paulo (USP), Pirassunuga, S\~{a}o Paulo, Brazil} 

\author{Odemir Martinez Bruno}
              \email{bruno@ifsc.usp.br}
\affiliation{Scientific Computing Group, S\~ao Carlos Institute of Physics, University of S\~{a}o Paulo (USP),  cx 369 13560-970 S\~{a}o Carlos, S\~{a}o Paulo, Brazil - www.scg.ifsc.usp.br\\ phone:+55 16 3373 8728 / fax:+55 16 3372 2218}

\date{\today}

\begin{abstract}
The use of a rapid and accurate method in diagnosis and classification of species and/or cultivars of forage has practical relevance, scientific and trade in various areas of study. Thus, leaf samples of fodder plant species \textit{Brachiaria} were previously identified, collected and scanned to be treated by means of artificial vision to make the database and be used in subsequent classifications. Forage crops used were: \textit{Brachiaria decumbens} cv. IPEAN; \textit{Brachiaria ruziziensis} Germain \& Evrard; \textit{Brachiaria Brizantha} (Hochst. ex. A. Rich.) Stapf; \textit{Brachiaria arrecta} (Hack.) Stent. and \textit{Brachiaria spp}. The images were analyzed by the fractal descriptors method, where a set of measures are obtained from the values of the fractal dimension at different scales. Therefore such values are used as inputs for a state-of-the-art classifier, the Support Vector Machine, which finally discriminates the images according to the respective species.
\end{abstract}


\maketitle

\section{Introduction}  
\label{sec:intro}

The knowledge and understanding of functional properties of plants make possible to develop advances in several areas like medicine to cure diseases, produce and improve species to feed people and animals \cite{Camargo2009}. Involving this last topic, the analysis of consumption by animals is very important because the animal production can be improved from grazed pastures.  Specifically, ruminants have their amount of feeding directly linked with the processes of particle-size reduction during the feeding. Due to the physical strength of grasses, ruminant animals consume larger quantities of forages with lower resistance to breakdown \cite{Herrero2001}. Since the grass is extremely important for animal food, it becomes the object of study here being one of the main forms of ruminant feeding is through grazing \textit{Brachiaria}.


The genus Brachiaria consists of herbaceous, perennial or annual, erect or decumbent. Belonging to the grass family, it presents approximately one hundred species, and therefore their correct classification is of great importance for the genetic improvement of forage species and purity of the species in the field of seed production \cite{P72,WCMRN00,ATTBP13}. Grouping plants into genus is a way of facilitating the understanding of the diversity of the grasses, according to the particularities of each species \cite{P72}.


The classification of grasses is mainly based on the characters of the spikelet structure and its arrangement. The main taxonomic feature of the  genus \textit{Brachiaria}, despite not being present in many species, is the reversed or adaxial position of the spikelet and of ligule. This spikelet is relatively large, oval or oblong and it is arranged regularly in a row along one side of the rachis. However, the taxonomy of this genus is not satisfactory, both in terms of species composition and in their inter-relationship with other genus. Problems related to incorrect classifications often occur among Brachiaria species commonly used in pastures, as well as among accessions of germplasm collections.


Since there is a great variability among natural species of \textit{Brachiaria}, to identify really discriminant characters becomes a difficult task so that seeking for techniques that improve the identification will contribute to studies within this theme, as well as provide a reasonable system of classification, since there is no such system for the genus \textit{Brachiaria}. Thus, the objective of this study was to take the volumetric Bouligand-Minkowski and Probability fractal descriptors associated with the Principal Component Analysis transform to classify samples from five species of \textit{Brachiaria} cultivars. This methodology provides a set of coefficients for each image that will characterize it. The tests were performed in a large database with almost ten thousand of samples including the superior and inferior face of leaves obtaining 92.84\% of correctness rate in the classification (of all leaves).

The text in this paper is organized as follows. In Sections \ref{sec:fractalGeometry}, \ref{sec:probability} and  \ref{sec:fractalDescriptors} the theory of the methods is explained. In Section \ref{sec:method} the description of the method Bouligand-Minkowski with probability dimension applyed to texture characterization. Section \ref{sec:experiments} shows the experiments in a database of  \textit{Brachiaria} leaves and in Section \ref{sec:results} the results are analysed. The paper is concluded in Section \ref{sec:conclusions}.

\section{Fractal Geometry}
\label{sec:fractalGeometry}

Fractal geometry \cite{M68} is the area of Mathematics which deals with fractal objects. These are gometrical structures characterized by two main properties: the infinite self-similarity and infinite complexity. In other words, these elements are recursively composed by similar structures. In addition, they exhibit a high level of detail on arbitrarily small scales.

In the same way as in the Euclidean geometry, fractal objects are described by numerical measures. The most widespread of such measures is the fractal dimension. Given a geometrical set $X$ (set of points in the $N$-dimensional space), the fractal dimension of $D(X)$ is expressed in the following equation:
\[
	D(X) = N - \lim_{\epsilon \rightarrow 0}\frac{\log(\mathfrak{M}(\epsilon))}{\log(\epsilon)},
\]
where $\mathfrak{M}$ is a fractality measure and $\epsilon$ is the scale parameter. The literature presents various definitions for the fractality measure \cite{T95,R94}. The following sections describe two of such approaches.

\subsection{Bouligand-Minkowski}

One of the best-known methods for estimating the fractal dimension of an object is the Bouligand-Minkowski approach \cite{T95}. In this solution, the grayscale image $I:[1:M] \times [1:N]$ is mapped onto a surface $S$, using the following relation:
\[
	S = \{(i,j,k)|(i,j) \in [1:M]\times[1:N],k = I(i,j)\}.
\]
Then, each point having co-ordinates $(x,y,z)$ is dilated by a sphere with variable radius $r$. Therefore, the dilation volume $V(r)$ may be computed by the following expression:
\[
	V(r) = \sum{\chi_{\mathfrak{D}(r)}[(i,j,k)]},
\]
where $(i,j,k)$ are points in the surface $S$, $\chi$ is the characteristic function and $\mathfrak{D}(r)$ refers to the following set:
\[
	\mathfrak{D}(r) = \{(x,y,z)|[(x-P_x)^2 + (y-P_y)^2 + (z-P_z)^2]^{1/2} \leq r\},
\]
in which $(P_x,P_y,P_z) \in S$. In practice, the Euclidean Distance Transform \cite{FCTB08} is used to determine the value of $V(r)$.

\begin{figure}[!htd] 
\centering
\mbox{\subfigure[]{\includegraphics[width=0.3\columnwidth]{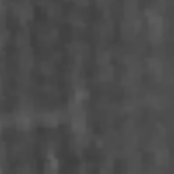}}\hspace{10mm}
\subfigure[]{\includegraphics[width=0.45\columnwidth]{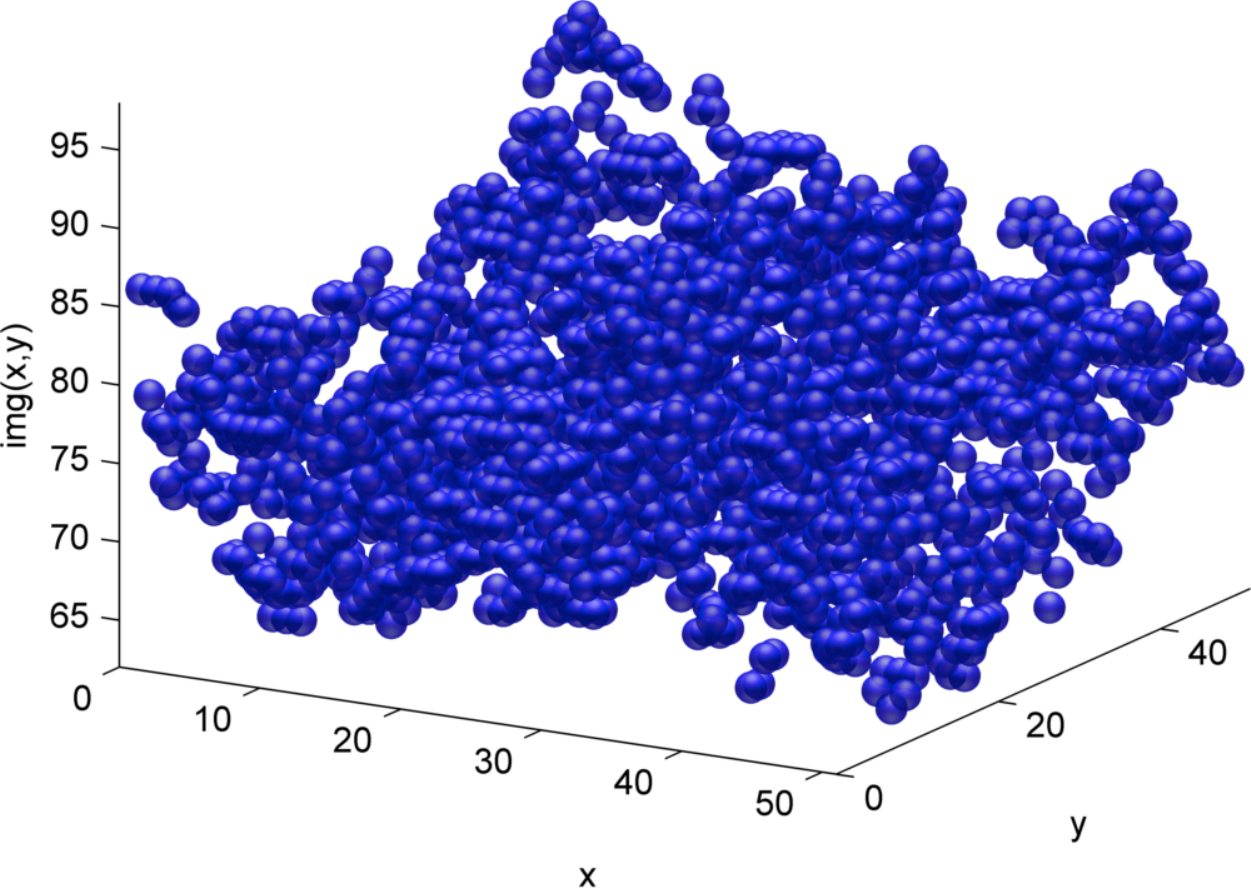}} }
\mbox{
\subfigure[]{\includegraphics[width=0.45\columnwidth]{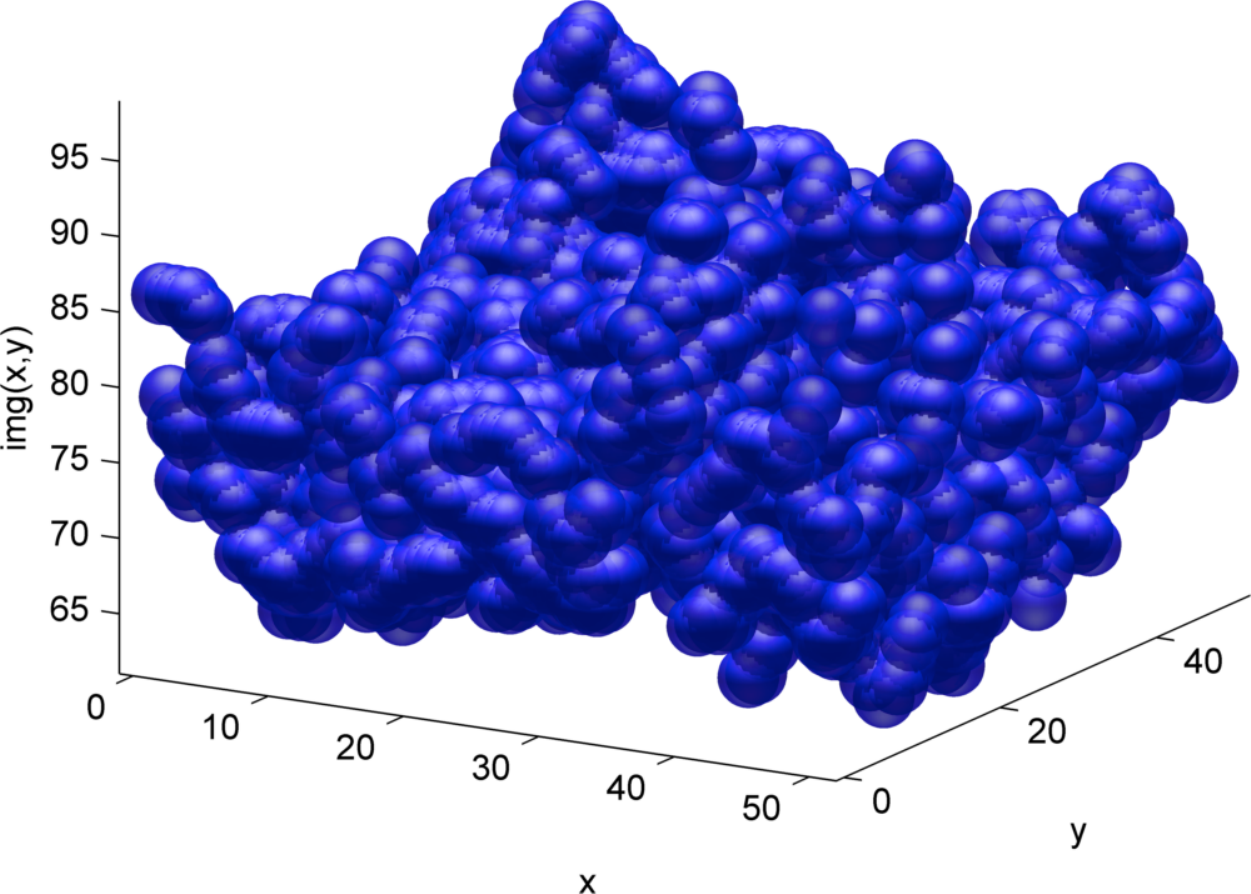}}
\subfigure[]{\includegraphics[width=0.45\columnwidth]{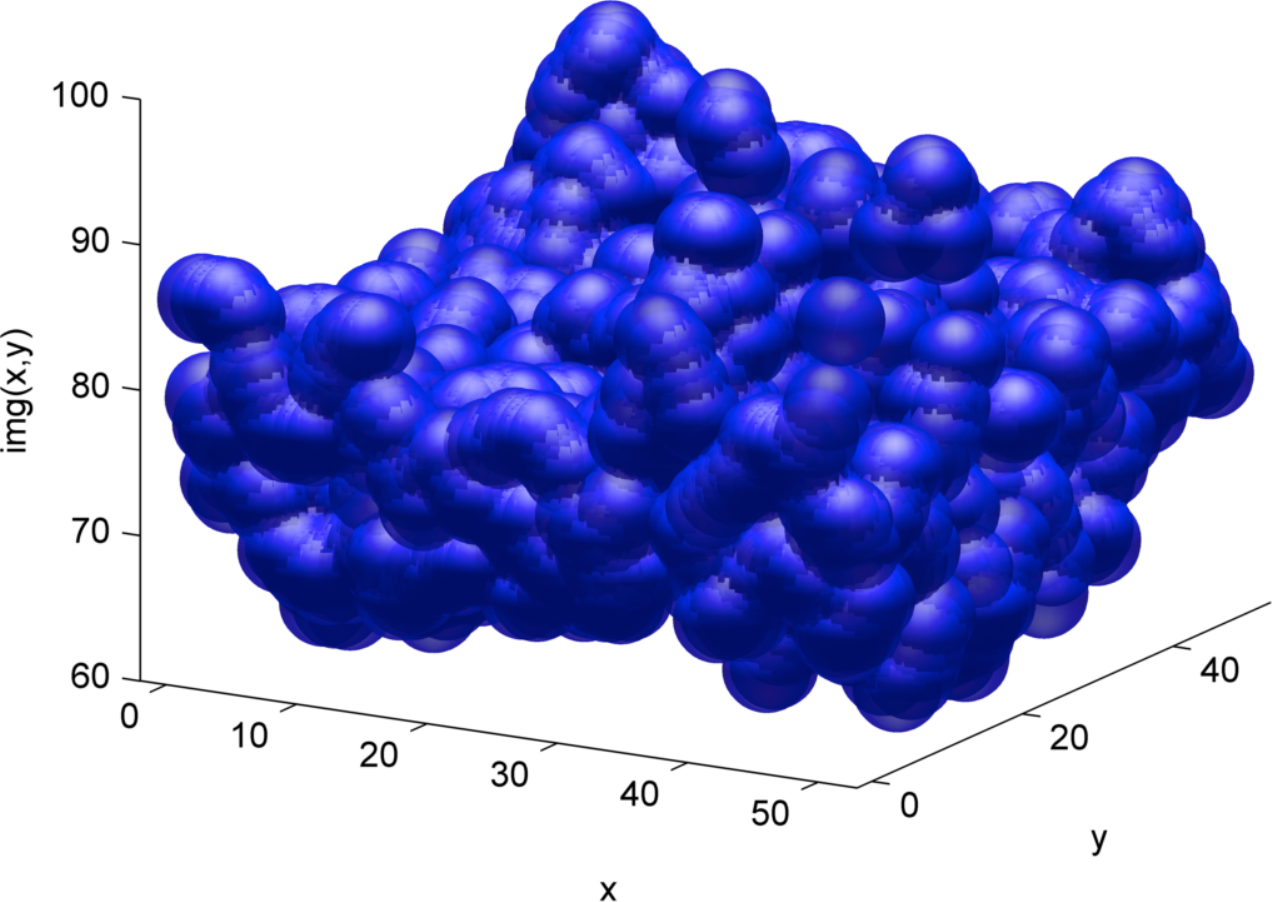}}}
\caption{Bouligand-Minkowski dilation process. a) Gray-level image. b) Surface with radius 1. c) Surface with radius 2. d) Surface with radius 3.}
\label{fig:mink}
\end{figure}

Finally, the fractal dimension $D_{BM}$ itself is given by:
\[
	D_{BM} = 3 - \lim_{r \rightarrow 0}\frac{\log(V(r))}{\log(r)}.
\]
The limit in the above expression is calculated by plotting the values of $V(r)$ against $r$, in $\log-\log$ scale, and the limit is the slope of a straight line fitting the $\log-\log$ curve. The Figure \ref{fig:mink} exemplify the process.

\section{Probability Dimension}
\label{sec:probability}

Also refered to as Voss dimension, this method obtains the fractal dimension from the statistical distribution of pixel intensities within the image \cite{V86}.

Like in the Bouligand-Minkowski method, the image analyzed is converted into a three-dimensional surface $S$. Hence, the surface is surrounded by a grid of cubes with side $\delta$. By varying the value of $\delta$, the information function $N_P$ is provided through:
\[
	N_P(\delta) = \sum_{m=1}^{N}{\frac{1}{m}p_m(\delta)},
\]
where $N$ is the maximum possible number of points within a single cube and $p_m$ is the probability of $m$ points in $S$ belonging to the same cube.

Finally, the fractal dimension $D$ is estimated by the following relation:
\begin{equation}\label{eq:prob}
	D = -\lim_{\delta \rightarrow 0}\frac{\ln N_P}{\ln \delta}.
\end{equation}
As with the Bouligand-Minkowski approach, the limit is computed by a least squares fit.

\section{Fractal Descriptors}
\label{sec:fractalDescriptors}

Fractal descriptors are a methodology that extracts meaningful information of an object of interest by means of an extension of the fractal dimension definition \cite{BPFC08,FB12b,FCB11}. Instead of using the dimension for describing the object, the fractal descriptors $u$ use the entire set of values in the fractality curve:
\begin{equation}
	u:\log(\epsilon) \rightarrow \log(\mathfrak{M}).
\end{equation}
The values of $u$ can be used directly to compose the feature vector \cite{FB12b} or after some sort of transform \cite{FCB11}. In the present work, we use the raw data of $\log(\mathfrak{M}$.

The fractal descriptors extract significant information from the object at different scales. The values of the fractality for larger radii measure the global aspect of the structure. In the case of plant leaves like those analyzed here, these data concern important information regarding the general aspect of the nervure distribution. On the other hand, the smallest radii provide essential information on the variability of the pixel intensities inside a local neighborhood. Biologically, these micro-patterns are tightly related to the constitution of the plant tissue.

\section{Proposed Method}
\label{sec:method}

The proposed methodology combines the previously described approaches to compose a precise and robust tool to identify plant species from an image of the leaf of \textit{Brachiaria}. The feature vector is obtained by concatenating the Bouligand-Minkowski and Probability descriptors and then applying a dimensionality reduction procedure to the merged descriptors.

Let the Bouligand-Minkowski descriptors be represented by the vector $\vec{D_1}$:
\[
	\vec{D_1} = \{x_1, x_2, ...,x_{n_1}\}
\]
and the Probaility features expressed through $\vec{D_2}$:
\[
	\vec{D_2} = \{y_1, y_2, ...,y_{n_1}\}.
\]
Then, in a problem of species discrimination over a database of leaf images, we can define two feature matrices, $M^{(1)}_{m \times n_1}$ and $M^{(2)}_{m \times n_2}$, one for each descriptor, where the rows correspond to the descriptor vectors for each image to be analyzed. After that, the matrices are concatenated horizontally, giving rise to the matrix $M$. Next, this matrix is transformed into one matrix $\tilde{M}$:
\begin{equation}
	\tilde{M} = M_{inter} M_{intra}^{-1},
\end{equation}
where $M_{intra}$ and $M_{inter}$ are, respectively, the inter and intra-class matrix. The intra-class matrix is defined by:
\begin{equation}
	S_{intra} = \sum_{i = 1}^{K}{\sum_{i \in C_i}{(X(i,.)-\overline{C_i})(X(i,.)-\overline{C_i})^T}},
\end{equation}
where $C_i$ is the $i^{th}$ class in $M$, $K$ is the total number of classes, $M(i,.)$ expresses the $i^{th}$ row (sample) of $M$ and $\overline{C_i}$ is a row vector representing the average descriptors  of each class $C_i$. On its turn, the inter-class matrix is provided by the following expression:
\begin{equation}
	S_{inter} = \sum_{i=1}^{K}{N_i(\overline{C_i}-\overline{M})(\overline{C_i}-\overline{M})^T},
\end{equation}
in which $N_i$ is the number of samples of the $i^{th}$ class.

The concatenation process ensures that the resulting descriptors emphasize the best discriminative properties of each fractal approach. In this case, both descriptors provide a different perspective of the object. While the Bouligand-Minkowski capture a multiscale mapping of the texture morphology, the Probability method gives a detailed description of the statistical distribution of the pixels along the gray-level image. The sum of these viewpoints makes possible a detailed description of the patterns within the image, at different scales.

\begin{figure*}[!htpb]
	\centering
		\includegraphics[width=0.7\textwidth]{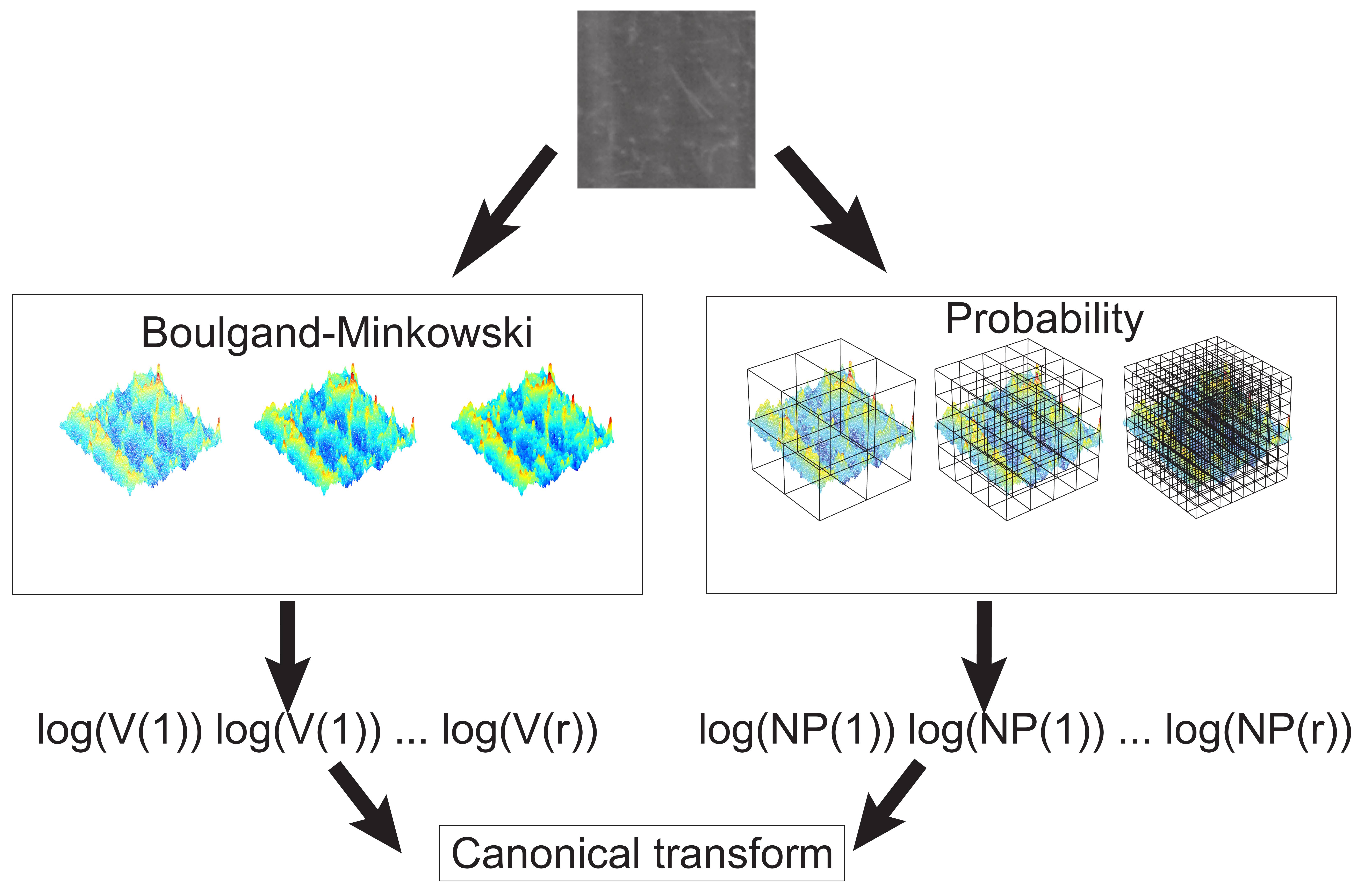}
	\caption{A diagram illustrating the steps to obtain the proposed descriptors.}
	\label{fig:method}
\end{figure*}

In the present study, the cancatenated descriptors are obtained from the windows extracted from the scanned image of the analyzed grass. The Figure \ref{fig:method} illustrates the process involved in computing the descriptors, since the original image until the descriptors themselves.

\section{Experiments} 
\label{sec:experiments}

The leaf samples were collected in the agrostologic field at the Faculdade de Zootecnia e Engenharia de Alimentos (FZEA-USP), which is located in Pirassununga city at state of S\~ao Paulo, Brazil. The leaf images were collected manually, directly from live plants, with extreme carefulness in not damaging the leaf surface. All plants grew with ideal conditions of nutrients and lighting. In this study, five species were took: 
\textit{Brachiaria decumbens} Stapf. cv. Ipean 
\textit{Brachiaria ruziziensis} Germain $\&$ Evrard, 
\textit{Brachiaria brizantha} (Hochst. ex. A. Rich.) Stapf., 
\textit{Brachiaria arrecta} (Hack.) Stent. 
and \textit{Brachiaria spp.}. 

After collecting the leaves, they were submitted to a scanning procedure, where the superior and inferior face of the leaf was scanned in 1200 dpi (\textit{dots-per-inch}) resolution and saved in a lossless image format with no compression. It was obtained 5 sheets with 10 different tillers, totalling 100 samples (50 images from the superior faces of the leaves and 50 images from the inferior faces).

\begin{figure*}
\center
\includegraphics[width=0.5\textwidth]{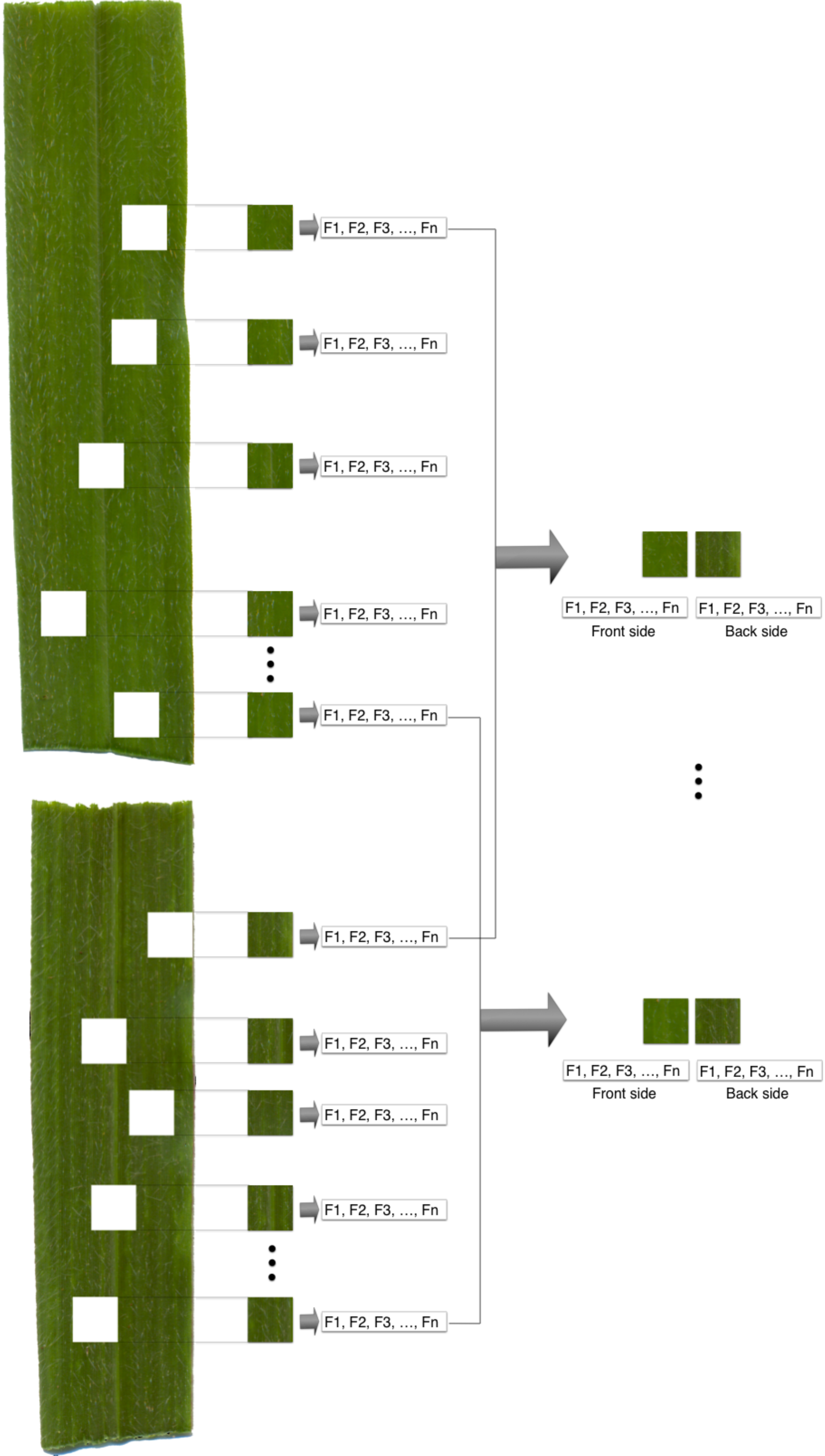}
\caption{\label{fig:janelas}Samples acquisition. Sub-images of 200 $\times$ 200 pixels size obtained from samples. A signature $F = \{F1, F2, F3, ..., Fn$\} is extracted for each sub-image and, afterward, the signature from the superior and inferior face of the leaf are concatenated to obtain the final signature.}
\end{figure*}

\begin{figure*}
\centerline{
\subfigure[]{
\includegraphics[width=7 cm]{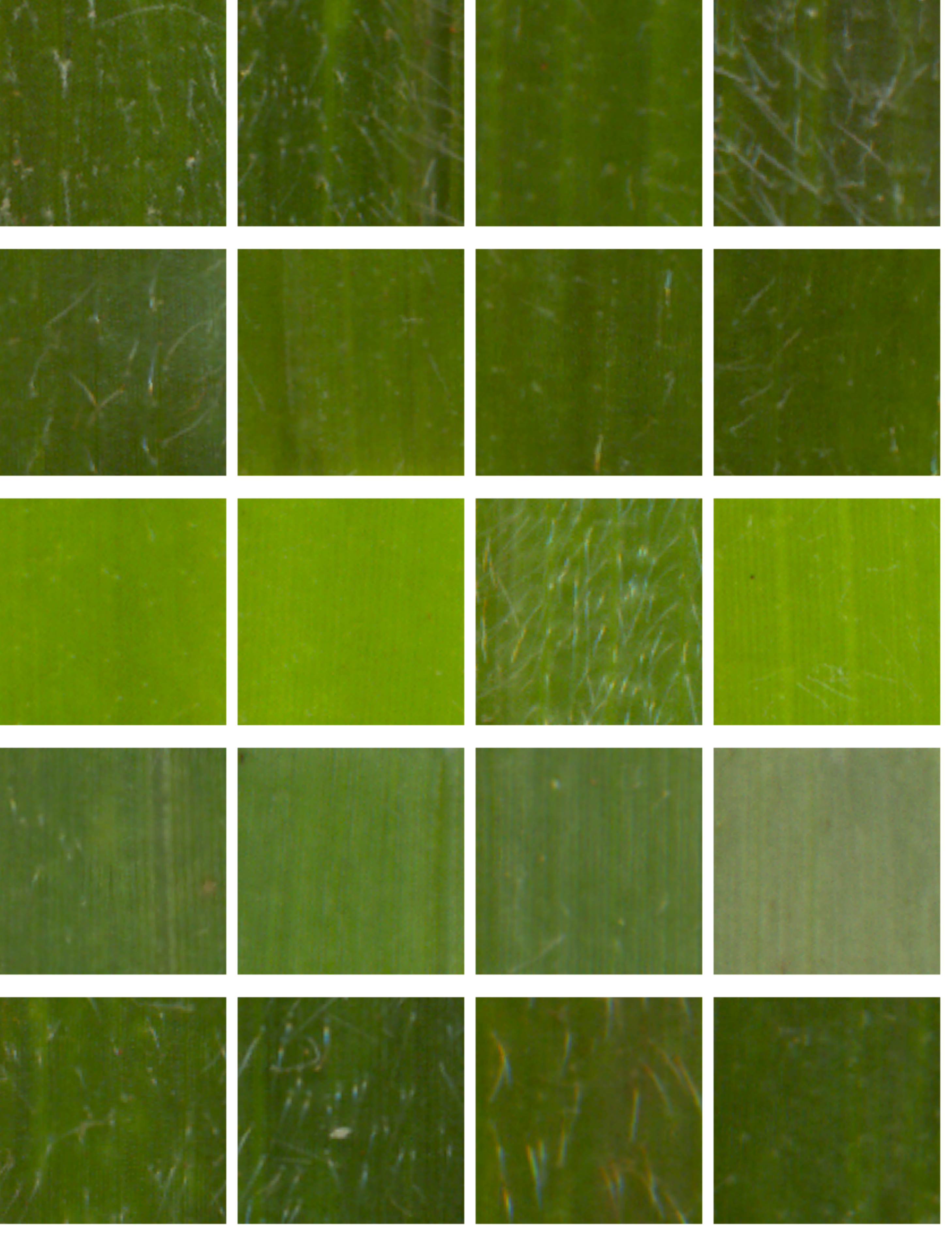}
\label{fig:janelasfrente}}
\hfil
\subfigure[]{
\includegraphics[width=7cm]{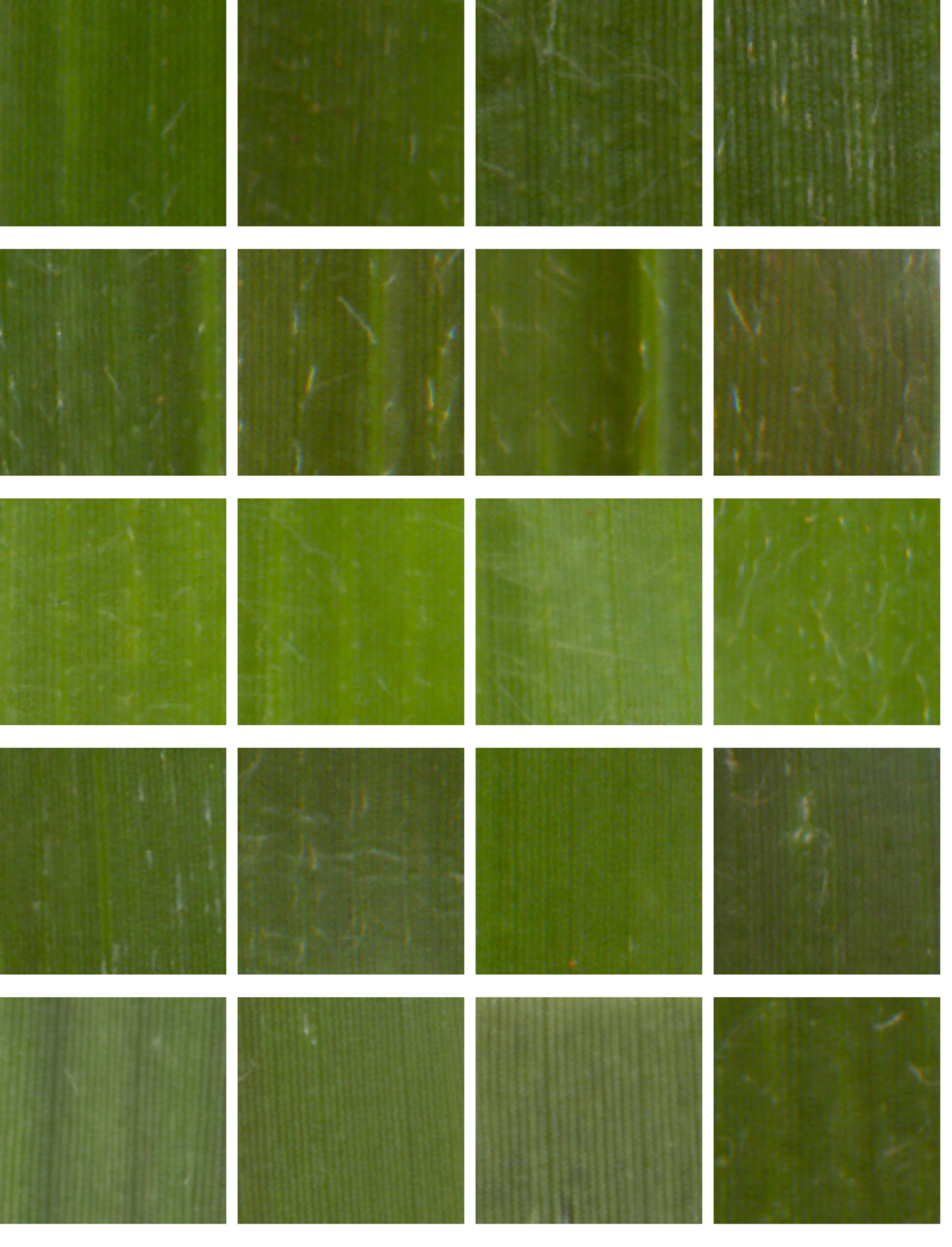}
\label{fig:janelasverso}}}
\caption{\label{fig:imgsdatabase}Samples of final database. Each row represents samples of the same specie, from top to bottom: \textit{Brachiaria decumbens}, \textit{Brachiaria ruziziensis}, \textit{Brachiaria brizantha}, \textit{Brachiaria arrecta} and \textit{Brachiaria spp.}. (a) Face superior  and (b) face inferior of  leaf.}
\end{figure*}

As the leaves were scanned manually, they were not properly aligned. Therefore, the images were vetically aligned according to the central axis using the Radon transform \cite{Deans1993}. Subsequently, for mounting the database, it were randomly obtained about 20 sub-images of 200 $\times$ 200 pixels without overlapping and considering all the leaf surface, avoiding stains margins and allowing the central vein (Figure \ref{fig:janelas}). At Figure \ref{fig:janelas} it can also be seen the preparation of the signature for each sample. The method is applied to each sub-image from the sample (superior and inferior face of the leaf) to obtain a feature vector $F$ with features \{$F1, F2, F3, ..., Fn$\} where $n$ is the number of features.  Afterward, the signature from the superior face of the image is concatenaded with the signature from the inferior face to generate the final signature of the sample. Therefore, the final database is composed by 9832 images, 4916 images from superior face and 4916 from inferior face of leaves. Figure \ref{fig:imgsdatabase} shows some samples for each class of the database.


The proposed method is applied to compute descriptors from the images of the analyzed grass species. Therefore, the obtained descriptors are used as the input of a classifier, in this case, the Suport Vector Machine method \cite{V99}. The classification is carried out in a 10-fold cross-validation scheme \cite{V99}.

\section{Results}
\label{sec:results}

\begin{figure}[!htpb]
	\centering
		\includegraphics[width=0.5\textwidth]{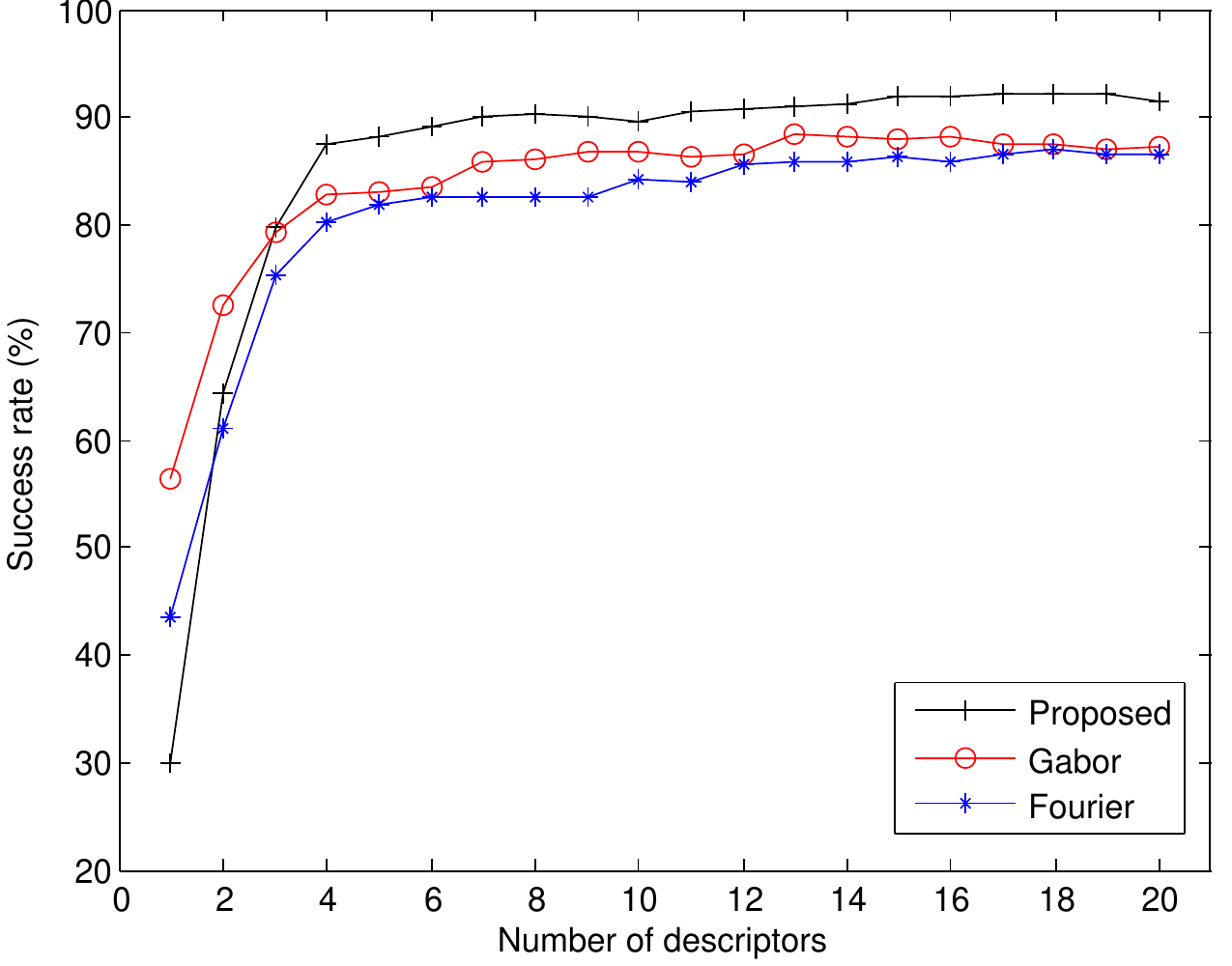}
	\caption{Graph of the success rate as function of the number of descriptors.}
	\label{fig:acerto}
\end{figure}

The graph in the Figure \ref{fig:acerto} shows the success rate of each compared texture descriptors when the number of descriptors is varied. We notice that even with only 4 elements, the proposed method achieves a correctness rate close to 90\%. Gabor and Fourier have a close behavior, while the fractal descriptors presents an outstanding performance for any number of elements greater than 3.

The Table \ref{tab:success} shows the best success rate achieved by each method by using an optimal number of descriptors. The success rate is accompanied by other important measures related to the performance of each approach applied to discriminate the grass species. In this table, $ND$ is the number of descriptors, $CR$ is the correctness rate, $\kappa$ is the $\kappa$ index and $AE1$ and $AE2$ are the type 1 and 2 errors. The number of descriptors ranges from 1 to 30, as after this point the performance tends to stabilise. The greatest success rate and smallest errors were obtained by the combined fractal descriptors. Such outcome confirms the effectiveness of the proposed methodology in this classification task, providing an excellent categorization of the plant species.
\begin{table}[!htpb]
	\centering
		\begin{tabular}{cccccccc}
			\hline
                 Method & ND & CR (\%) & $\kappa$ & AE1 & AE2\\
                 \hline
                 \hline
                  Fourier & 30 & 89.79 & 0.88 & 0.10 & 0.10\\
                  Gabor & 21 & 88.02 & 0.86 & 0.12 & 0.12\\
                  Proposed method & 30 & 92.84 & 0.91 & 0.07 & 0.07\\
                  \hline			
		\end{tabular}
	\caption{Success rate for different approaches.}
	\label{tab:success}
\end{table}

The tables \ref{tab:CM_prop}, \ref{tab:CM_fourier} and \ref{tab:CM_gabor} exhibit the confusion matrices for each compared descriptor. These tables are helpful to describe either the correctly classified samples as well as the false negatives and false positives, outside the main diagonal. We observe that, despite some minor differences among the methods, the methodology developed in the present study demonstrated to be the most reliable solution to identify the grass samples. Reinforcing the values on the Table \ref{tab:success}, the type 1 and 2 errors are very similar, expressing the homogeneity of the distinguished classes.
\begin{table}[!htpb]
	\centering
		\begin{tabular}{ccccc}
         \hline
          913 & 7 & 20 & 6 & 14\\
          12 & 888 & 19 & 8 & 73\\
          25 & 15 & 925 & 6 & 5\\
          3 & 14 & 2 & 952 & 9\\
          20 & 80 & 4 & 10 & 886\\          
          \hline			
		\end{tabular}
	\caption{Confusion matrix for the proposed method.}
	\label{tab:CM_prop}
\end{table}
\begin{table}[!htpb]
	\centering
		\begin{tabular}{ccccc}
         \hline
           882 & 21 & 22 & 2 & 33\\
           20 & 884 & 22 & 16 & 58\\
           22 & 24 & 877 & 34 & 19\\
           14 & 20 & 21 & 919 & 6\\
           52 & 74 & 12 & 10 & 852\\          
          \hline			
		\end{tabular}
	\caption{Confusion matrix for the Fourier method.}
	\label{tab:CM_fourier}
\end{table}
\begin{table}[!htpb]
	\centering
		\begin{tabular}{ccccc}
         \hline
           891 & 35 & 2 & 9 & 23\\
           33 & 829 & 13 & 59 & 66\\
           0 & 2 & 958 & 15 & 1\\
           40 & 55 & 19 & 813 & 53\\
           37 & 102 & 5 & 20 & 836\\          
          \hline			
		\end{tabular}
	\caption{Confusion matrix for the Gabor method.}
	\label{tab:CM_gabor}
\end{table}

The great performance achieved by the proposed fractal descriptors is due to the nature of the fractal modeling, as mathematical fractals and natural objects have a great deal in common. The similarities are related to the high complexity usually found in the nature as well as the self-similarity property, which also appears often in parts of plants, like leaves, flowers, etc. Actually, the fractal descriptors are tightly related to important physical attributes of the leaf, such as roughness, reflectance and distribution of colors and brightness levels. In turn, this set of properties is capable of identify plant species faithfully, using their digital image representation, as demonstrated in \cite{BPFC08}. The present study confirms such reliability and robustness of fractal descriptors and shows that this is a powerful tool for the species categorization of the grasses analyzed here.

\section{Conclusions}
\label{sec:conclusions}

The present study proposed a combination of fractal descriptor approaches to discriminate among species of \textit{Brachiaria} grass, based on the digital images from their leaves. The proposed solution achieved a high success rate even using a low number of features. Such result confirms the effectiveness and reliability of fractal descriptors in this kind of task.

This result is also remarkable from a biological perspective, as \textit{Brachiaria} grasses are one of the most important foods for animals that are used for the labor and human consumption. The precise discrimination of species makes possible to better understand the distribution of species in a region and, as a consequence, to optimize the necessary attention for that region.

Despite the importance of this study, the literature shows very few works on \textit{Brachiaria} classification and the present is the first to obtain such a great effectiveness. Such so good results suggest the use of fractal descriptors as a powerfull method to identify these species and markedly help the taxonomy specialist.

\section*{Acknowledgments}
 
O. M. Bruno gratefully acknowledges the financial support of CNPq (National Council for Scientific and Technological Development, Brazil) (Grant Nos. 308449/2010-0 and 473893/2010-0) and FAPESP (Grant No. 2011/01523-1). 
N. R. da Silva acknowledges support from FAPESP (The State of S\~ao Paulo Research Foundation). 
J. F. Batista acknowledges support from FAPESP (The State of S\~ao Paulo Research Foundation).


\end{document}